\documentclass[letterpaper]{article} 
\usepackage{aaai2026}  
\usepackage{times}  
\usepackage{helvet}  
\usepackage{courier}  
\usepackage[hyphens]{url}  
\usepackage{graphicx} 
\usepackage{natbib}  
\usepackage{caption} 
\usepackage{algorithm}
\usepackage{algorithmic}
\usepackage{amsmath}
\usepackage{multirow} 
\usepackage{booktabs}
\usepackage{amssymb}  
\usepackage{xcolor}   

\usepackage{newfloat}
\usepackage{listings}

\lstdefinelanguage{yaml}{
  keywords={true,false,null,y,n},
  keywordstyle=\color{blue}\bfseries,
  basicstyle=\ttfamily\small,
  comment=[l]{\#},
  commentstyle=\color{gray}\ttfamily,
  stringstyle=\color{orange},
  moredelim=[l][\color{black}]{:},
  morestring=[b]',
  morestring=[b]"
}

\DeclareCaptionStyle{ruled}{labelfont=normalfont,labelsep=colon,strut=off} 
\floatstyle{ruled}
\newfloat{listing}{tb}{lst}{}
\floatname{listing}{Listing}
\title{D-GARA: A Dynamic Benchmarking Framework for GUI Agent Robustness in Real-World Anomalies}
\author{
  Sen Chen\textsuperscript{\rm 1}\equalcontrib,
  Tong Zhao\textsuperscript{\rm 1}\equalcontrib,
  Yi Bin\textsuperscript{\rm 1}\thanks{Corresponding author.},
  Fei Ma\textsuperscript{\rm 2},
  Wenqi Shao\textsuperscript{\rm 3},
  Zheng Wang\textsuperscript{\rm 1}
}
\affiliations{
  \textsuperscript{\rm 1}Tongji University\\
  \textsuperscript{\rm 2}Guangdong Laboratory of Artificial Intelligence and Digital Economy (SZ)\\
  \textsuperscript{\rm 3}Shanghai AI Laboratory\\
  cs.sen@hotmail.com, zt.tong@hotmail.com, yi.bin@hotmail.com
}
\begin{document}

\maketitle

\begin{abstract}
Developing intelligent agents capable of operating a wide range of Graphical User Interfaces (GUIs) with human-level proficiency is a key milestone on the path toward Artificial General Intelligence. While most existing datasets and benchmarks for training and evaluating GUI agents are static and idealized, failing to reflect the complexity and unpredictability of real-world environments, particularly the presence of anomalies. To bridge this research gap, we propose D-GARA, a dynamic benchmarking framework, to evaluate Android GUI agent robustness in real-world anomalies. D-GARA introduces a diverse set of real-world anomalies that GUI agents commonly face in practice, including interruptions such as permission dialogs, battery warnings, and update prompts. Based on D-GARA framework, we construct and annotate a benchmark featuring commonly used Android applications with embedded anomalies to support broader community research. Comprehensive experiments and results demonstrate substantial performance degradation in state-of-the-art GUI agents when exposed to anomaly-rich environments, highlighting the need for robustness-aware learning. D-GARA is modular and extensible, supporting the seamless integration of new tasks, anomaly types, and interaction scenarios to meet specific evaluation goals. 
\end{abstract}

\begin{links}
    \link{Code}{https://github.com/sen0609/D-GARA}
    \link{Extended version}{https://sen0609.github.io/D-GARA/}
\end{links}

\section{Introduction}
A GUI agent is designed to operate a wide variety of Graphical User Interfaces (GUIs), simulating human interaction through visual interface understanding, task planning, and action execution. This capability represents a crucial milestone toward the vision of Artificial General Intelligence. In early approaches, GUI agents relied on predefined templates, empirical rules, or heuristic search trees to accomplish simple tasks, which are inflexible and difficult to generalize to real-world scenarios. Powered by Vision-Language Models (VLMs), recent GUI agents have made significant progress in executing complex tasks, demonstrating strong visual understanding and task planning abilities, particularly on a range of static benchmarks. However, such success in ideal environments often obscures their underlying fragility in dynamic and realistic settings. Real-world tasks are inherently more challenging, and execution environments are more complex, usually involving unpredictable events and interruptions, which significantly limits their practical deployment and applicability.

\begin{figure}[t]
    \centering
    \includegraphics[width=0.47\textwidth]{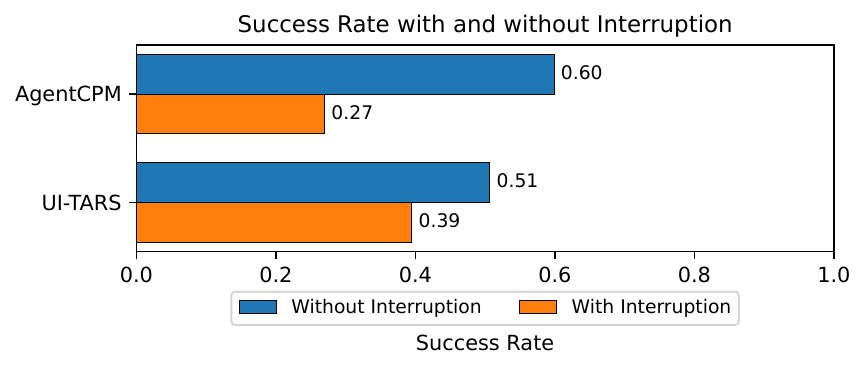}
    \caption{Comparison of task success rates for UI-TARS and AgentCPM under normal conditions and real-world interruptions.}
    \label{fig:sr_decline}
\end{figure}

However, current GUI agents largely ignore such interruptions and are typically trained on idealized datasets, \textit{i.e.}, sequences of screenshots paired with corresponding actions to accomplish given tasks. To quantitatively investigate the impact of real-world interruptions on GUI agents, we evaluate two representative state-of-the-art models specifically designed for GUI interaction: UI-TARS-72B and AgentCPM-GUI-8B. As shown in Figure~\ref{fig:sr_decline}, both models experience a significant decline in task success rate when subjected to real-world anomalies, with performance degradation reaching up to 33\%. Obviously, existing static and idealized benchmarks, \textit{e.g.}, Android Control ~\cite{li2024effectsdatascaleui}, fail to expose the vulnerability of GUI agents or to assess their robustness in the presence of interruptions. In practice, application interfaces are dynamic, and actions of an agent may trigger unforeseen system events which would lead to different ways to reach the task goal. Although recent work has begun to introduce ``anomalies'' into evaluation settings~\citep{yang2025guirobustcomprehensivedatasettesting}, these efforts remain confined to static paradigms and fall short of simulating sudden and process-altering events that arise during real-time interactions. AndroidWorld~\cite{rawles2025androidworlddynamicbenchmarkingenvironment} provides a dynamic benchmarking environment while the included applications and tasks are pure and simple, failing to simulate the complex real-world operation process. Therefore, a standardized framework is urgently needed to comprehensively evaluate agent robustness against anomalies in dynamic and real-world environments.

To fill this research gap, we propose to design a novel \textbf{D}ynamic benchmarking framework for \textbf{G}UI \textbf{A}gent \textbf{R}obustness in real-world \textbf{A}nomalies, dubbed \textbf{D-GARA}, tailored for Android GUI agents. The core idea of D-GARA is to transform the evaluation setting from static screenshots to a dynamic environment to simulates real-world interactions. Specifically, D-GARA integrates an Android simulator to construct a realistic Android environment for executing tasks. While a GUI agent performs tasks in this environment, D-GARA injects a series of high-frequency anomalies through a controllable injection system, including unexpected modal pop-ups, disruptive system alerts, and intermittent application crashes. Although GUI Robust~\cite{yang2025guirobustcomprehensivedatasettesting} also provides a small portion of anomalous samples, its static nature makes it only capable of simulating simple anomalies by inserting a screenshot of an alert or pop-up into a normal task trajectory. These anomalies are easy to simulate and straightforward for agents to bypass. Agents can simply click through them and continue along the original execution path. Even in the case of a mistake, the benchmark provides a gold-standard screenshot for the next action prediction, essentially following a teacher-forcing paradigm similar to next-token prediction in language models. In contrast to these static settings, D-GARA  steps further by introducing more complex and realistic anomalies that may lead to entirely different execution trajectories. For example, in real-world scenarios, certain anomalies may redirect the agent to unexpected screens outside the intended task flow. If the agent does not actively maintain awareness of the task goal, it can easily become misled by these unanticipated diversions. Such anomalies pose a significantly greater threat to task completion, as they can completely derail the agent from its intended plan. D-GARA is designed to simulate these severe anomalies, providing a more comprehensive evaluation of robustness and decision-making capabilities for GUI agent.

Besides, D-GARA facilitates the collection and annotation of dynamic benchmarking samples for Android GUI agents. It provides a dedicated tool named \verb|datacollector|, which assists users in gathering human trajectories, including screenshots and XML files—for their tasks, while offering the flexibility to inject interruptions at any desired point. We also include several common interruptions by default, such as Low Battery Dialogs and Location Permission. Moreover, users can easily define and configure custom interruptions for specific Apps or to meet particular requirements. Leveraging these capabilities, we contribute a benchmark collected from several commonly used applications, each embedded with diverse interruptions, to benefit the research community. In addition, we propose a robustness metric to evaluate how well GUI agents perform when exposed to interruptions, and we assess several state-of-the-art agents to identify their limitations and potential areas for improvement.

In summary, the main contributions of this paper are as follows:
\begin{itemize}

    \item We first explicitly consider interruptions during the real-world execution of GUI agents. To simulate such scenarios, we introduce D-GARA, a novel benchmarking framework for assessing GUI agent robustness under real-world anomalies, which also supports dynamic evaluation and real-world data collection.

    \item We collect and annotate one of the first benchmarks involving commonly used applications, intentionally triggering a wide range of real-world interruptions to examine how agents respond to unexpected events. This benchmark will be open-sourced to support the future research in the GUI agent community.

    \item We conduct extensive experiments on state-of-the-art GUI agents using D-GARA, aiming to comprehensively evaluate the impact of anomalies and assess the robustness of agents when faced with such interruptions. We also propose a robustness metric for this purpose. Our experimental results show that all current SOTA methods suffer from significant performance degradation, highlighting the urgent need for more robust solutions.
\end{itemize}

\section{Related Work}

\subsection{GUI Agent}

The rapid development of Multimodal Large Language Models (MLLMs) has made the GUI agent paradigm both feasible and increasingly popular.These models can be broadly categorized into two types: closed-source models, such as GPT-4 ~\cite{openai2024gpt4technicalreport}, Gemini 2.5 ~\cite{geminiteam2025geminifamilyhighlycapable} and open-source models, including Qwen2.5-VL~\cite{bai2025qwen25vltechnicalreport}, InternVL ~\cite{chen2024internvlscalingvisionfoundation} and MiniCPM~\cite{yao2024minicpmvgpt4vlevelmllm}. 

Closed-source models typically exhibit superior general-purpose reasoning and planning capabilities, enabling them to better understand complex tasks and follow instructions. As a result, many GUI agent systems leverage these models as planners while relying on separate modules for grounding. Notable examples include GUI Actor ~\cite{wu2025guiactorcoordinatefreevisualgrounding} and U-Ground ~\cite{gou2025navigatingdigitalworldhumans}, both of which use powerful language models for high-level decision making while incorporating specialized grounding mechanisms to interact with the interface.

To address privacy concerns from handling sensitive data, recent work favors compact open-source models for local deployment, aiming to improve grounding and planning while ensuring practical usability.
For example, UI-TARS-1.5~\cite{qin2025ui}, an open-source multimodal agent built on top of Qwen2.5-VL~\cite{bai2025qwen25vltechnicalreport}, which incorporates reinforcement learning to enhance planning and reasoning capabilities during inference. Another representative model is AgentCPM-GUI~\cite{zhang2025agentcpmgui}, developed by THUNLP, Renmin University of China, and ModelBest. It is optimized for on-device deployment and is built upon the lightweight MiniCPM-V model~\cite{yao2024minicpmvgpt4vlevelmllm}. With reinforcement fine-tuning and pre-training on a large bilingual Android dataset named CAGUI~\cite{zhang2025agentcpmgui}, it excels in grounding Chinese mobile UI elements and executing real-world tasks across more than 30 popular applications. 
These works reflect a growing emphasis on building GUI agents that are not only capable of understanding user interfaces, but also practical for real-world deployment through targeted post-training and efficient architectures.

\subsection{GUI Benchmark}

Benchmark datasets are essential for evaluating GUI agents, and can be broadly divided into two categories based on the core agent capabilities they target: grounding benchmarks and task-completion benchmarks. Grounding-focused benchmarks, such as ScreenSpot~\cite{cheng2024seeclick}, ScreenSpot-v2~\cite{wu2024osatlasfoundationactionmodel} and ScreenSpot-Pro~\cite{li2025screenspotpro}, evaluate an agent’s ability to localize UI elements within static screenshots. Task-completion benchmarks, such as
Mind2Web (Deng et al. 2023) and GUI-Odyssey (Lu
et al. 2024a), assess the agent’s capability to perform
multi-step interactions and complete realistic tasks using
screenshots. Recently, several benchmarks have been proposed
for dynamic GUI environments, such as Android
World ~\cite{rawles2025androidworlddynamicbenchmarkingenvironment} is a dynamic benchmarking environment which provide 116 tasks,but the problem is it is most of its tasks is pure and simple,and it don't support researchers to add their own tasks. and also it is a anomaly-free benchmark.
Recent efforts such as GUI-Robust~\cite{yang2025guirobustcomprehensivedatasettesting} exposes agent limitations under irregular GUI states but relies on static overlays (e.g. pop-ups, login pages) that are shallow and reversible. Real-world anomalies are dynamic and non-deterministic, often requiring real-time recovery and replanning.
To overcome this, we propose D-GARA, a dynamic framework that injects anomalies during live Android task execution. It enables realistic, high-fidelity robustness evaluation beyond what static benchmarks can offer.

\begin{figure*}[t]
    \centering
    \includegraphics[width=0.9\textwidth]{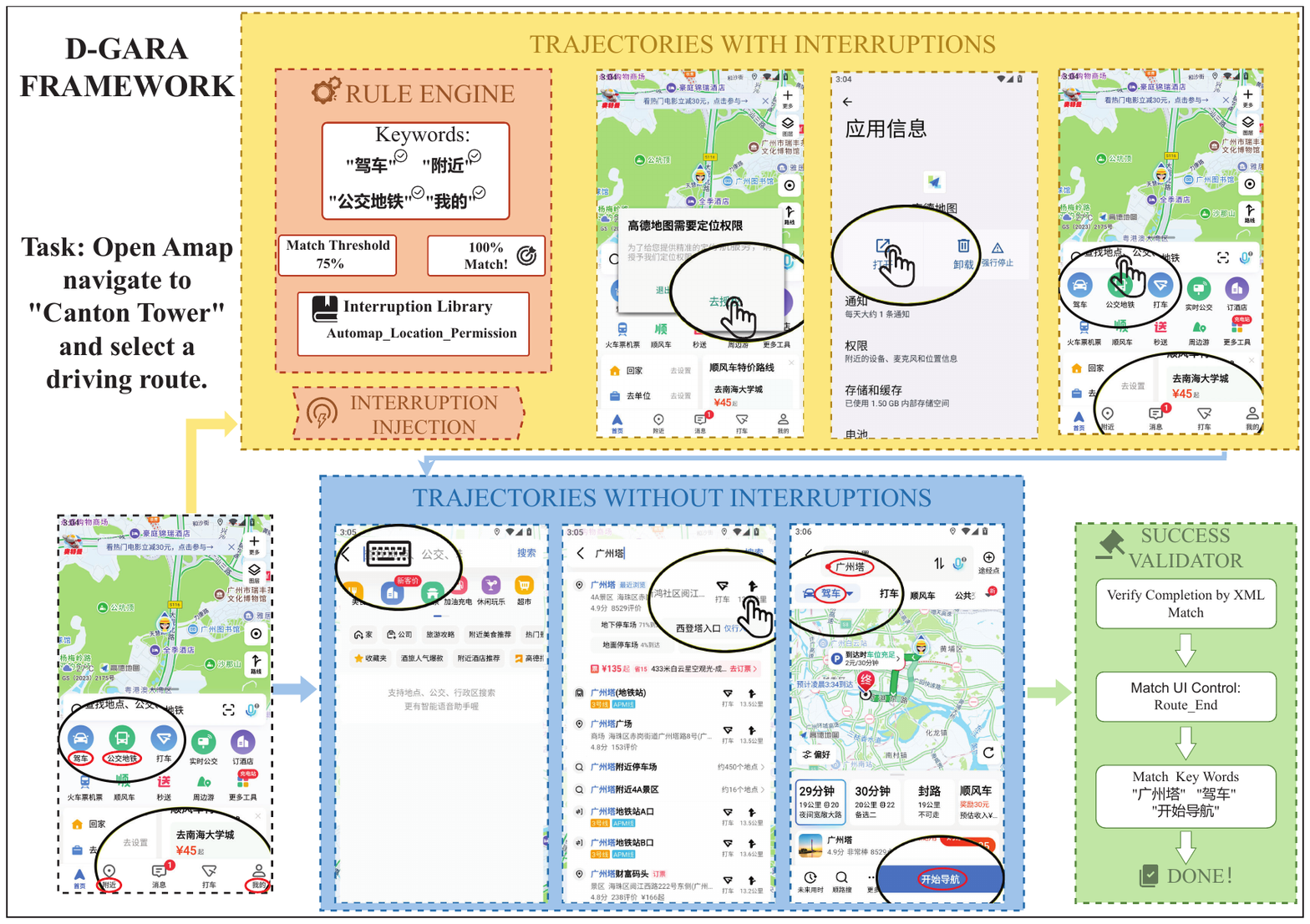}
    \caption{Pipeline of D-GARA Framework for Real-World GUI Agent Robustness Evaluation, Including Rule-Based Interruption Injection and Goal-State Validation}
    \label{fig:framework_overview}
\end{figure*}

\section{The D-GARA Framework}
\label{sec:framework}

D-GARA incorporates interruption injection and success validation into the execution trajectory of agent to better simulate real-world disturbances and evaluate task completion. Figure \ref{fig:framework_overview} provides an overview of the framework, and the following section details each component.

\subsection{Framework Overview}

As shown in Figure~\ref{fig:framework_overview}, D-GARA is a modular execution framework that coordinates agent interaction, real-time anomaly injection, and success validation. The process begins by capturing a screenshot of the initial state and the corresponding UI layout (XML hierarchy) from the Android device. The framework then decides whether to inject an interruption based on the current state. If triggered, a new screenshot and XML file are collected and passed to the agent. With this multimodal input, the agent produces an action command, e.g., a tap or text entry, which is sent to the device via \texttt{ADB}. After a brief pause for the UI to stabilize, the framework evaluates the updated UI state. The above process forms an ``execution cycle'' in which two core modules operate sequentially: 1) \textit{the anomaly trigger mechanism}, which determines whether a context-aware interruption is needed, and 2) \textit{the success validation mechanism}, which checks whether the agent has reached the goal. Together, these mechanisms support D-GARA’s integrated interruption triggering and success validation.

\subsection{Interruption Trigger Mechanism}
To make the interruption injection more natural and reasonable, we propose \textbf{Semantic Anomaly Triggering Mechanism}, which inspects the textual content of the XML file for specified keywords using a lightweight rule-based evaluator. Each rule defines a target condition through a set of keywords and an associated matching threshold. An anomaly is triggered only when the current XML file meets all required conditions. For example, as shown in Figure~\ref{fig:framework_overview}, a navigation-related rule may specify keywords such as Drive, Nearby, Metro, and Mine (translated from Chinese) in the UI, and it is triggered when the XML file contains at least three of these terms. Once the condition is met, the framework injects a predefined interruption, such as a permission dialog, system alert, or other interruptions. A concrete rule specification for this case is shown in Listing~\ref{lst:drive_rule}.

\begin{listing}[ht]
\caption{Anomaly injection rule for the Drive page.}
\label{lst:drive_rule}
\begin{lstlisting}[language=yaml]
- id: rule_drive_permission
  conditions:
    all:
      - type: semantic_element_exists
        keywords: ["Drive", "Nearby", "Metro", "Mine"]
        threshold: 0.75
  actions:
    - type: inject_interference
      interference_id: "location_permission_dialog"
      follow_up: 
        accept: "redirect_to_settings"
        deny: "terminate_app"
\end{lstlisting}
\end{listing}

Building on this basic trigger mechanism, we design a two-stage anomaly handling pipeline that models both the interruption pop-up displayed on the screen and the follow-up actions after the agent responds. In Stage 1, the agent detects and handles a real dialog in the foreground, such as a location permission prompt. In Stage 2, D-GARA uses ADB commands to trigger the follow-up action based on the agent’s choice, simulating the dynamic state transition and executing the corresponding system action. For example, when opening Amap, the system may display a location permission dialog with two options: Accept or Deny. If the agent accepts, the workflow redirects to the settings interface to enable the permission. If denied, the application terminates, since it cannot work without the location permission. These follow-up outcomes illustrate how a single interaction can lead to different execution paths, making anomaly handling more realistic compared to static pop-ups.

To support flexible anomaly definitions, D-GARA places all trigger logic in external configuration files instead of hard coding it. Each anomaly rule prototype specifies the target activity, the triggering condition, the content to display, and the follow up actions. This separation makes the logic easy to modify while keeping the framework stable. Researchers can create or adjust anomaly scenarios by editing the configuration files, providing a modular and practical foundation for evaluating agent robustness.

\subsection{Success Validation Mechanism}

Traditional evaluations of GUI agents on static datasets assume a fixed action sequence: each screenshot corresponds to a specific action, and the task concludes once all frames are processed. In dynamic environments, however, these assumptions no longer apply, because actions may not be unique, states can evolve unpredictably, and there is often no clear signal for task completion or goal achievement. While the action space typically includes a "done" option to indicate task completion, this signal is unreliable. Small models may fail to produce it even after completing the task, while larger models may misuse it and terminate prematurely. As a result, relying on model-generated signals does not guarantee consistent task verification across different models or task types. These limitations highlight the need for a state-centered validation strategy, where completion is judged based on the actual UI state rather than the agent’s self-reported signals.

To handle these issues, D-GARA introduces the \textit{Success Validator}, a state centered evaluation module that verifies progress at every state update. After each agent action, the framework records a new screenshot and the corresponding XML view hierarchy. Before passing this state back to the agent, the validator inspects the XML content to determine whether it satisfies a declaratively defined goal condition. Listing~\ref{lst:success_rule} provides an example rule for a video related task. In this case, the validator checks whether the property of a specific UI element contains the value ``Liked'', which indicates that the task has been successfully completed.

\begin{listing}[t]
\caption{A success validation rule for a video-like task.}
\label{lst:success_rule}
\begin{lstlisting}[language=yaml]
task_drive_like_video:
  conditions:
    - type: element_property_contains
      selector: "resource-id=tv.app:id/like_button"
      attribute: "content-desc"
      value: "Liked"
\end{lstlisting}
\end{listing}

In general, the validator performs this check after every agent action. When the agent explicitly outputs a completion signal such as ``done'', the validator immediately verifies the result. If the goal condition is not satisfied at that moment, the task is stopped and counted as a failure.

A task is considered successful once the \textit{Success Validator} confirms that the stabilized UI state matches the specified goal schema. These schemas describe the expected final interface, such as the presence of particular elements or the values of their properties. With this design, task success depends only on the final interface state rather than on the correctness of every intermediate action. In practice, this allows an agent to take detours, make temporary mistakes, or backtrack, as long as it eventually reaches the required goal state. This outcome centered design ensures that task completion is evaluated consistently across applications and environments. As a final safeguard, all automatically validated results are reviewed by human evaluators to ensure reliability in cases where automatic rules may be insufficient. In future work, we will explore more to involve large foundation model for this verification.

\section{Benchmark Construction}

Building upon the D-GARA framework, we construct \textbf{D-GARA-152}, a benchmark that includes 152 real world tasks across 8 widely used Android applications. While Section~\ref{sec:framework} introduced the general execution framework, this section explains how D-GARA is instantiated as a concrete benchmark through task selection, interruption design, goal specification, and evaluation metrics.

\subsection{Benchmark Overview}

D-GARA-152 focuses on applications that exhibit functional complexity, high user activity, and diverse interaction patterns, including e-commerce (e.g., JD.com, Amazon), social media (e.g., Weibo, Facebook), content consumption (e.g., Bilibili), navigation (e.g., Amap, Google Maps), and travel services (e.g., Ctrip). Task distribution is intentionally non-uniform, prioritizing high usage platforms to better reflect realistic agent workloads.

To emulate real-world disturbance patterns, each task is paired with one or more injected interruptions drawn from several representative categories. The interruption classification and injection strategy are detailed in Section~\ref{sec:interruption_strategy}. This strategy provides broad coverage of interruption types and keeps the scenarios aligned with real usage patterns.

\subsection{Interruption Design Strategy}
\label{sec:interruption_strategy}

Our interruption classification consists of five categories that cover common disruption sources in mobile environments:
\begin{itemize}
\item \textbf{System Resource}: low-battery warnings, thermal throttling alerts
\item \textbf{System Network}: Wi-Fi disconnection, mobile data switches
\item \textbf{App Malfunction}: crashes, freezes
\item \textbf{Permission Control}: runtime permission dialogs
\item \textbf{UX Disruption}: update prompts, feedback forms
\end{itemize}

Interruption injection follows a template based approach. We design general dialog layouts in Android Studio and compile them into a standalone APK. At runtime, D-GARA fills these templates with values defined in the external configuration files, which allows a single layout to support many interruption scenarios. Using a dedicated APK keeps interruption logic isolated from the target applications and avoids the inconsistencies of native system dialogs across devices and Android versions. Because both the layout and the textual content are parameterized, new interruptions can be added by extending the configuration library without modifying the core framework, keeping the mechanism portable, reusable, and easy to extend.

\subsection{Success Validation Design}

Building on the \texttt{Success Validation} module (Section~3.3), adapting it to the 152 benchmark tasks required a systematic process supported by human review.

We used a lightweight \texttt{DataCollector} tool to record the target screenshot and XML file for each application after a human operator completed a representative task. The designer then examined this final XML file to identify stable and unambiguous indicators of success. These XML attributes are usually more stable than visual appearance, because they reflect the underlying structure of the interface rather than transient graphical elements. Many applications use the same interface to mark completion for all tasks of a given type. For example, the rule in Listing~\ref{lst:success_rule} captures the success condition for a ``like'' action, and this condition applies to every task in that category for the same application. Once such indicators are extracted, this extracted success condition can be reused across all related tasks without additional manual effort. This reuse greatly reduces the cost of building large task suites, because the designer does not need to craft separate conditions for individual tasks. 

\subsection{Evaluation Metrics}

With task success defined through declarative rules, we next introduce the metrics used to evaluate agent performance under both baseline and interrupted conditions as follows:

\noindent\textbf{Success Rate (SR)} measures the proportion of tasks successfully completed under a given condition.  

\noindent\textbf{Robust Success Rate (RSR)} quantifies robustness over tasks solvable in baseline conditions:
\begin{equation*}
\text{RSR} = \frac{ \left| \left\{ i \mid SR_{\text{baseline}}^{(i)} = 1 \land SR_{\text{interruption}}^{(i)} = 1 \right\} \right| }{ \left| \left\{ i \mid SR_{\text{baseline}}^{(i)} = 1 \right\} \right| }.
\end{equation*}
RSR isolates interruption robustness from overall task-solving competence, enabling fair comparison among agents with different baseline capabilities. For instance, an agent may achieve a high SR yet exhibit a low RSR, indicating fragility under disturbances, whereas an agent with moderate SR but high RSR demonstrates stronger adaptability. Together, SR and RSR provide a comprehensive view of agent performance under real-world disruptions.

\section{Results and Analysis} 

To assess and analyze the robustness of GUI agents under real-world interruptions, we conduct extensive experiments and in-depth evaluations using several state-of-the-art GUI agents and advanced multimodal large language models (MLLMs), \textit{e.g.}, GPT-4o and Gemini2.5.

\subsection{Overall Performance Impact}

We first investigate the robustness of GUI agents, such as UI-TARS-1.5-72B~\cite{qin2025ui} and AgentCPM-GUI-8B~\cite{zhang2025agentcpmgui}, focusing on their ability to maintain task success in the presence of common interruptions and anomalies. In addition to such agents explicitly designed and trained for GUI tasks, we also evaluate several general MLLMs, including the powerful GPT-4o~\cite{openai2024gpt4technicalreport}, Gemini-2.5~\cite{geminiteam2025geminifamilyhighlycapable} , and Qwen2.5-VL-7B~\cite{bai2025qwen25vltechnicalreport}, as many agent frameworks adopt these models as their backbone, where they have demonstrated strong performance across a variety of tasks.

\begin{table}[ht]
\centering
\captionsetup{position=bottom} 
\setlength{\tabcolsep}{1.3mm}{
\begin{tabular*}{\linewidth}{@{\extracolsep{\fill}}lccc}
\hline
\textbf{Model} & \textbf{SR (NoInt)} & \textbf{SR (WithInt)} & \textbf{RSR} \\
\hline
Gemini2.5-flash & 80.26\% & 68.42\% & 73.77\% \\
GPT-4o & 69.08\% & 60.53\% & 66.67\% \\
Qwen2.5-VL-7B & 69.08\% & 46.05\% & 53.33\% \\
UI-TARS-1.5-72B & 50.66\% & 39.47\% & 48.05\% \\
AgentCPM-GUI-8B & 59.87\% & 26.97\% & 39.56\% \\
\hline
\end{tabular*}
}
\caption{Performance of models under interruption and non-interruption conditions}
\label{tab:overall_performance}
\end{table}

As the results shown in Table~\ref{tab:overall_performance}, all models exhibit a significant decrease in task success rates under interruption (comparison between the first two columns), with an average drop exceeding 17.5\%. This indicates that none of the agents can effectively handle unforeseen events during execution. These findings also support our hypothesis that performance on static benchmarks does not reliably translate to robustness in dynamic real-world conditions. We observe that larger models, such as GPT-4o and Gemini-2.5, demonstrate stronger robustness and  greater ability to withstand interruptions. This phenomenon may be attributed to their powerful planning capabilities, as most interruptions can be mitigated through effective planning, a skill in which larger models have been shown to excel in other domains, such as mathematical reasoning~\cite{shi2024math,lin2023sphinx}. From the RSR values in the last column, we observe that even AgentCPM-GUI and UI-TARS-72B, trained specifically for GUI interaction, still exhibit the weakest robustness when exposed to interruptions. This suggests that their training may primarily adapt the model to visually perceive the UI interface, while their planning ability remains heavily dependent on the underlying base model.

\subsection{Robustness across Interruption Types}

\begin{figure}[t]
    \centering
    \includegraphics[width=1\linewidth]{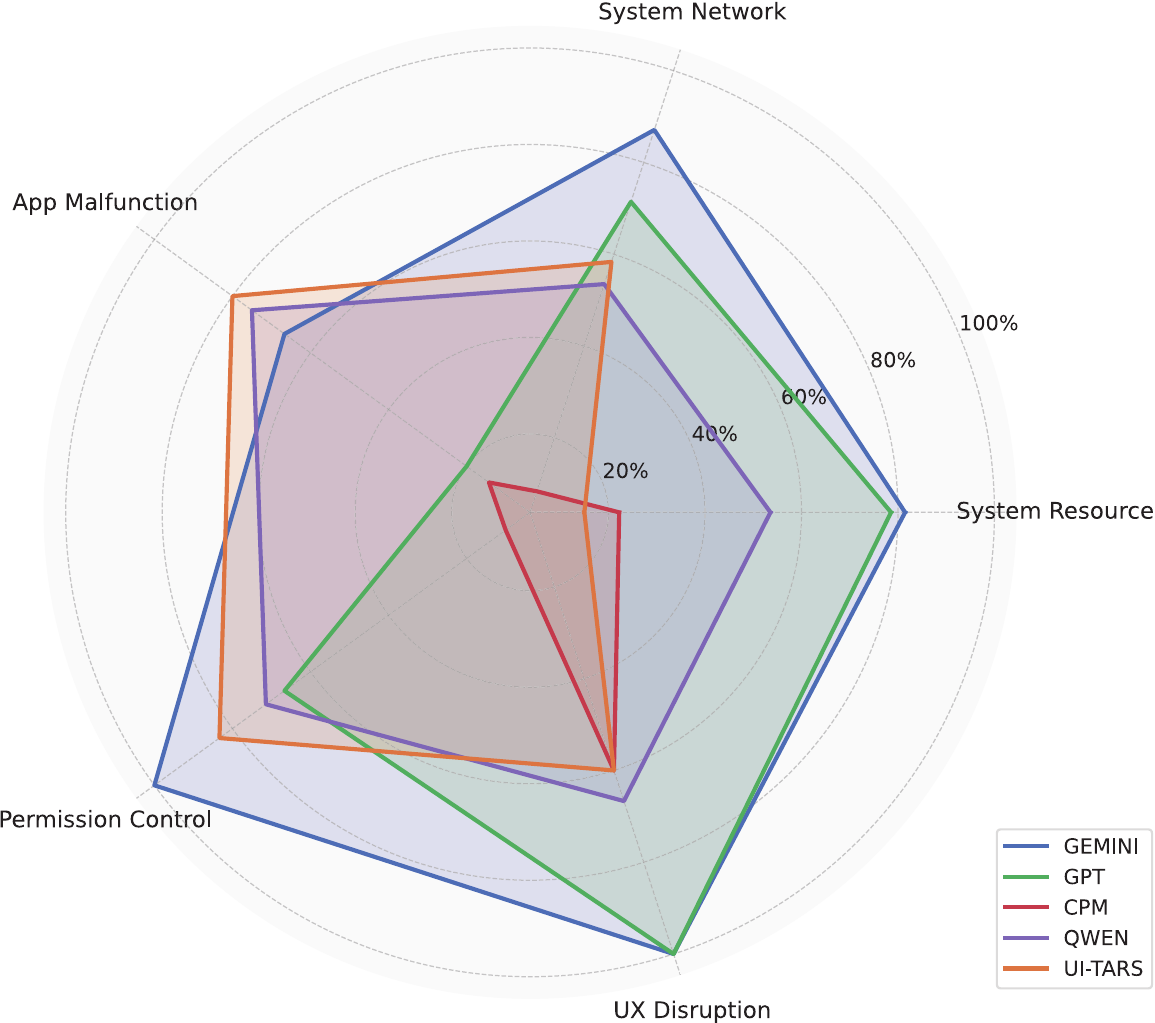}
    \caption{Robust success rate across five categories of GUI interruptions.}
    \label{fig:robustness_radar}
\end{figure}

As we know, different anomalies can lead to different execution paths, potentially disrupting perception, UI structure, or overall task flow. In other words, different types of interruptions vary in difficulty. To better analyze the effects of different interruption types, as aforementioned, we group them into five representative categories based on their functional characteristics and system scope: \textit{System Resource}, \textit{System Network}, \textit{App Malfunction}, \textit{Permission Control}, and \textit{UX Disruption}. Each category represents a distinct dimension of operational disruption, ranging from low-level system constraints to high-level user experience interruptions. The robust success rates (RSR) for each category are illustrated in Figure~\ref{fig:robustness_radar}. As we can see, Gemini and GPT-4o significantly outperform other methods on \textit{Permission Control} and \textit{System Network}, even achieving 100\% for \textit{Permission Control}, likely because these types of interruptions can be resolved with common knowledge, a strength of both models. For APP-level interruptions, which often require more specialized knowledge of UI interactions, UI-TARS achieves notably higher performance than in other interruption types, as it is trained for such tasks. Even AgentCPM demonstrates strong performance peak in this category, although it ranks lowest overall.

\begin{table}[ht]
\centering
\captionsetup{position=bottom}  
\small 
\setlength{\tabcolsep}{1.3mm}{
\begin{tabular}{lcc}
\hline
\textbf{Model} & \textbf{Dual-button RSR} & \textbf{Single-button RSR} \\
\hline
Qwen2.5-VL-7B & 96.15\% & 41.27\% \\
AgentCPM-GUI-8B & 82.35\% & 9.30\% \\
\hline
\end{tabular}
}
\caption{Robustness (RSR) comparison across models and interaction modes}
\label{tab:model_interaction_modes}
\end{table}

During our experiments, we observed that many interruptions present multiple response options, \textit{e.g.}, ``Install Now'' and ``Close'' in an APP update prompt, and agents typically choose ``Close'' to skip the interruption and continue the assigned task. Although this behavior demonstrates that agents can bypass interruptions, it makes us difficult to assess whether they can handle more complex paths. For example, selecting ``Install Now'' to complete the installation, and then return to the original task flow. To investigate this, we designed a specific test mode in which only the complex choice is available, \textit{e.g.}, ``Install Now'' here, forcing agents to follow the more challenging execution path. As shown in Table~\ref{tab:model_interaction_modes}, the performance drops sharply when agents are required to handle interruptions through the complex path. This finding suggests that agents need to strengthen their ability to truly manage complex interruptions, rather than simply skipping them by choosing ``Close''.

\subsection{The Effects of Action Coordinate}

Current GUI agents typically use both screenshots and \texttt{XML} element coordinates to ensure precise actions. While humans, as natural agents, rely solely on visual content to understand and execute tasks, as also applied in~\cite{lu2024omniparser}. To examine the impact of these modalities, we disentangle them and test two settings. In the \textit{screenshot-only} setting (as shown in Table~\ref{tab:modality_fusion}), agents receive only screenshots and are required to infer both the action and its coordinates. In the \textit{screenshot}+\texttt{XML} setting, agents predict the action from screenshots but obtained coordinates directly from the \texttt{XML} file. 

Results show that even the strong model Gemini2.5 performs poorly without \texttt{XML} input, exhibiting an almost 35\% drop in success rate, indicating that current agents struggle to infer accurate spatial coordinates from vision alone. In other words, the agents know \textit{what} action to take but not \textit{where} to execute it. In contrast, AgentCPM-GUI, which is trained specifically for GUI interaction, shows a smaller performance drop, suggesting it may have partially learned coordinate prediction. These findings highlight that enhancing visual perception and coordinate prediction could therefore be crucial to improving GUI agent performance.

\begin{table}[ht]
\centering
\captionsetup{position=bottom}  
\fontsize{9pt}{11pt}\selectfont
\setlength{\tabcolsep}{1.3mm}{
\begin{tabular}{llcc}
\hline
\textbf{Model} & \textbf{Modality}  & \textbf{SR (NoInt)} & \textbf{SR (WithInt)} \\
\hline
\multirow{2}{*}{AgentCPM-GUI} 
    & Screenshot+XML    & 59.87\% & 26.97\%\\
    & Screenshot-only   & 56.58\% & 19.74\%\\
\hline
\multirow{2}{*}{Gemini2.5-flash} 
    & Screenshot+XML    & 80.26\% & 68.42\%\\
    & Screenshot-only   & 45.33\% & 41.33\%\\
\hline
\end{tabular}
}
\caption{Comparison of AgentCPM-GUI and Gemini under modality settings}
\label{tab:modality_fusion}
\end{table}

\subsection{Perception Drift after App Crash}

\begin{figure}[ht]
    \centering
    \includegraphics[width=0.48\textwidth]{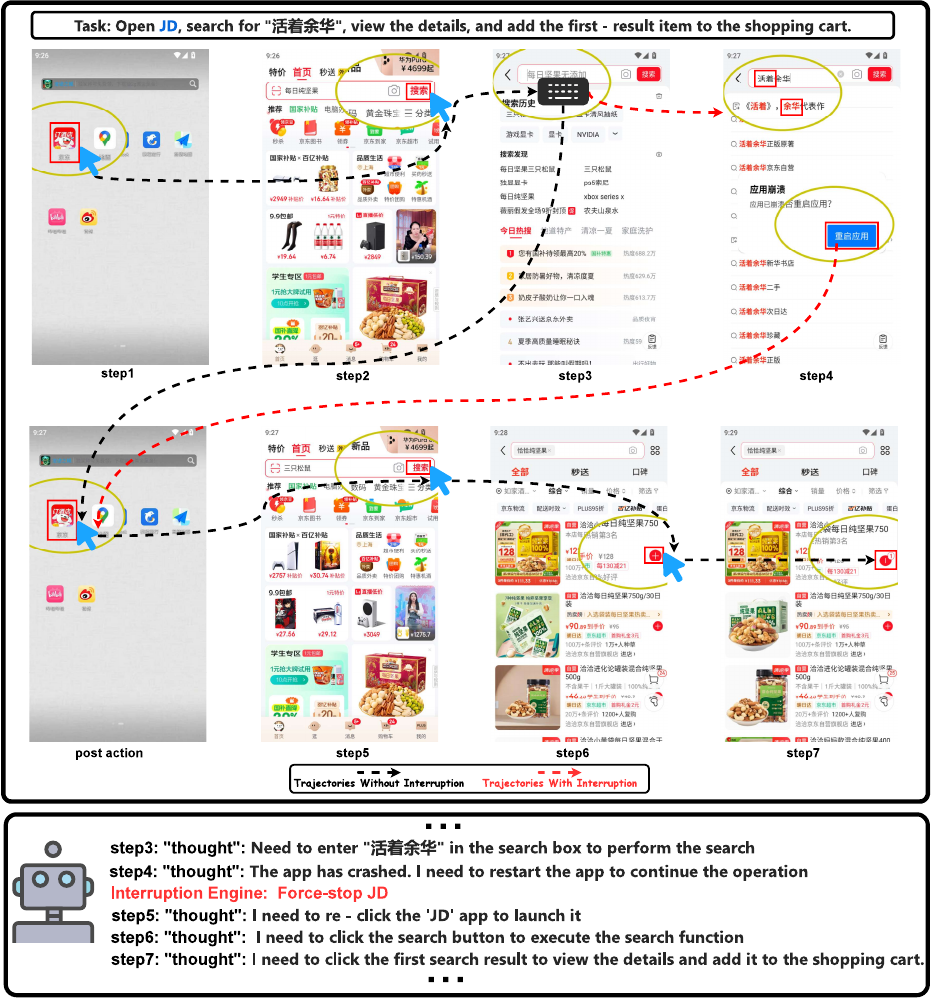} 
    \caption{A case of perceptual drift after App crash recovery. The black and red dash indicate normal and interruption path, respectively.}
    \label{fig:case_study}
\end{figure}

We observe an interesting phenomenon, in which GUI agents often fail to fully restart a task after encountering an app crash. As shown in Figure~\ref{fig:case_study}, recovering from a GUI anomaly is not the same as recovering from its disruptive effects, such as blocking actions or redirecting execution to unpredictable screens. Even after addressing the anomaly, for example by reopening the crashed app and returning to the previous page, residual effects may persist. In this case, GPT-4o correctly identifies the crash, relaunches the App, and appears to recover. However, upon returning to the search page, it deviates from the correct path, and then clicks the search button directly instead of re-entering the required search term, resulting in task failure.
This behavioral drift highlights a key issue that the decision-making of a model can be overly influenced by historical actions in the prompt, even when they no longer match the current visual state. In this case, GPT-4o has previously used the search interface, entered the target keyword, and logged that step in its action history. After the app crashes, upon returning to the same interface, it assumes that the keyword was still present and ignores that the search bar now contains recommended keywords. Improving the ability of an agent to retain useful memory while discarding misleading history could make it more intelligent and robust.

\section{Conclusion}
In this paper, we introduced D-GARA, a dynamic benchmarking framework designed to evaluate the robustness of GUI agents under real-world anomalies. Unlike prior static benchmarks that fall short in simulating the complexity of interactive environments, D-GARA enables active anomaly injection and state-centric validation in live Android settings. Our framework allows for flexible, extensible, and realistic robustness testing. Experimental results reveal significant performance degradation in existing state-of-the-art agents when exposed to dynamic interruptions, underscoring the need for robustness-aware learning and evaluation. By open-sourcing D-GARA, we hope to facilitate broader research into building more resilient and generalizable GUI agents that can operate effectively in unpredictable real-world environments. We encourage the community to build upon D-GARA by contributing new tasks, interruption types, and evaluation strategies to further advance robustness research in GUI agent development.

\bibliography{aaai2026}

\newpage
\section*{Appendix A: Benchmark Structure}

\begin{figure}[htbp]
    \centering
    \includegraphics[width=0.9\linewidth]{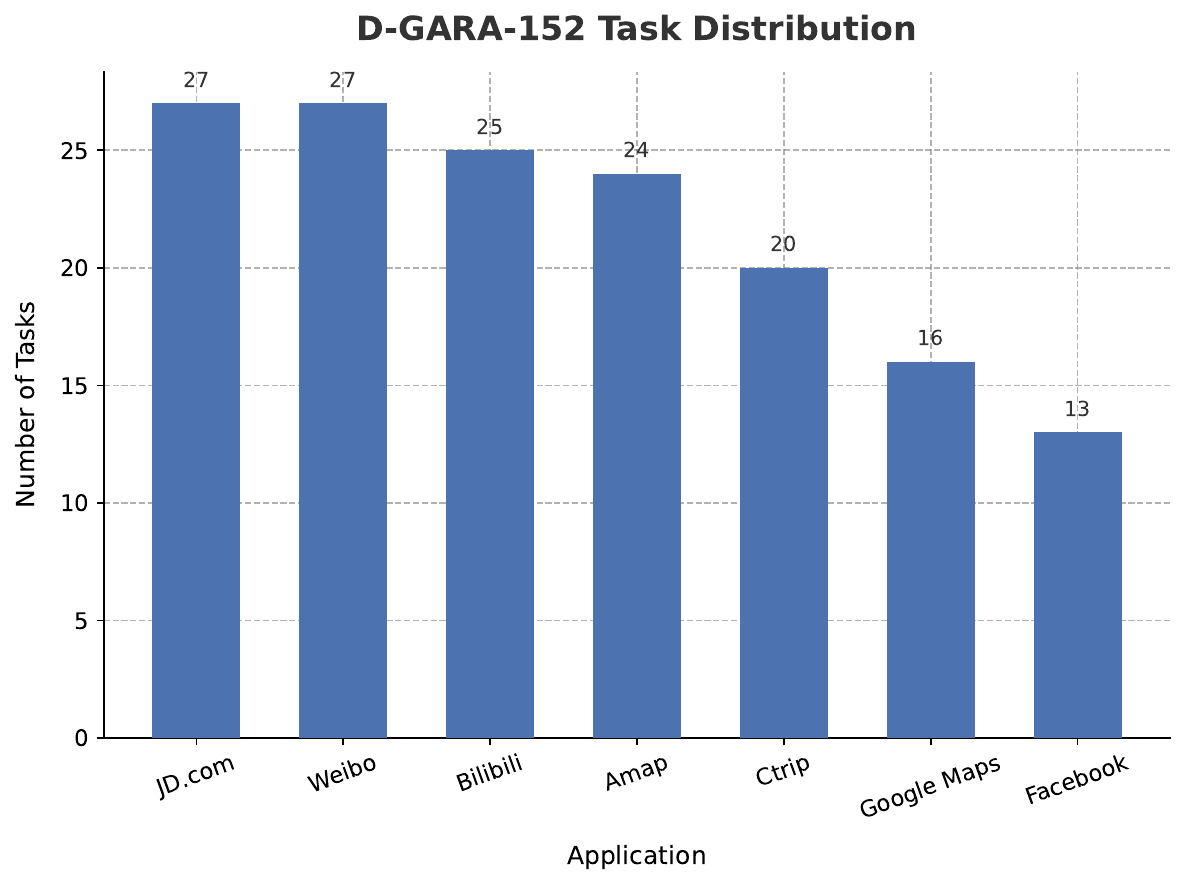}
    \caption{Distribution of tasks across different applications in D-GARA-152.}
    \label{fig:task_distribution}
\end{figure}

As shown in Figure ~\ref{fig:task_distribution}, the D-GARA-152 benchmark comprises 152 independent evaluation tasks distributed across 8 mainstream mobile applications. The task distribution is not uniform; instead, it strategically focuses on applications characterized by functional complexity, large user bases, and diverse interaction scenarios. This design aims to simulate the real-world challenges that GUI agents encounter. Regarding the task distribution, JD.com (27 tasks), Weibo (27 tasks), and Bilibili (25 tasks) are the three applications with the most tasks. This reflects the core of D-GARA-152 focus on e-commerce, social media, and content consumption. These platforms contain rich UI elements, varied user workflows, and frequent interface updates, making them ideal environments for testing agent robustness. Following them are Amap (24 tasks) and Ctrip (20 tasks), covering the map navigation and travel service domains. Additionally, Amazon (18 tasks) and Facebook (8 tasks) are included as international counterparts to ensure the generality of D-GARA-152 and cross-cultural adaptability. Finally, Google Maps (3 tasks) serves as a supplement to further diversify the navigation tasks.
\begin{figure}[htbp]
    \centering
    \includegraphics[width=0.9\linewidth]{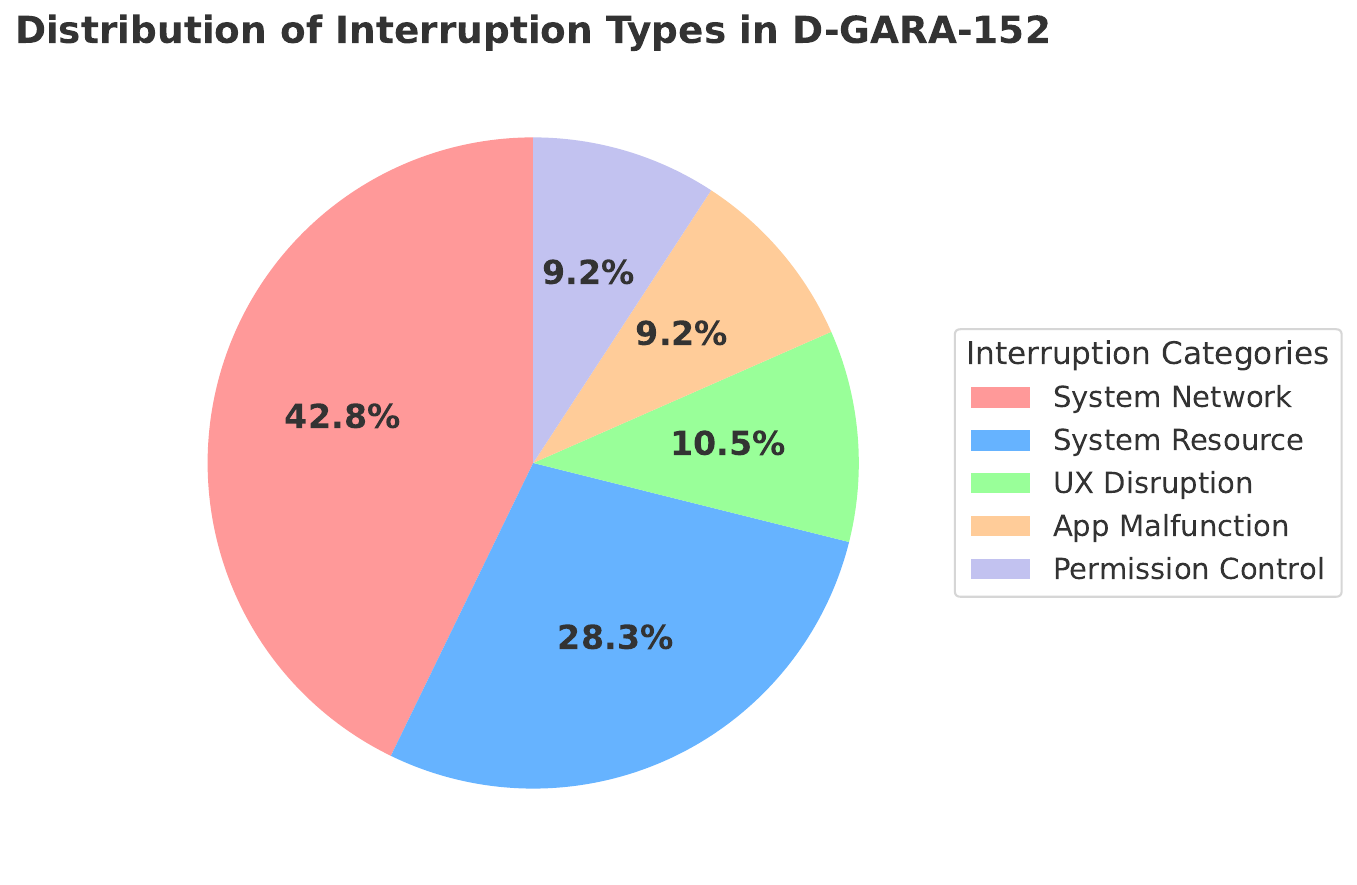}
    \caption{Distribution of interruption types in D-GARA-152.}
    \label{fig:interruption_distribution}
\end{figure}

In addition to task distribution across applications, the strength of the benchmark lies in its diverse and realistic interruption types. Following a structured taxonomy, we categorize these GUI interruptions into five distinct types. The distribution, detailed in Figure~\ref{fig:interruption_distribution}, is intentionally weighted to mirror real-world user experiences. System Network (42.8\%) and System Resource (28.3\%) interruptions constitute the vast majority of the benchmark, collectively accounting for over 70\% of all anomalies. This design choice is deliberate and reflects the reality that device-level issues, such as fluctuating network connectivity and low battery warnings, are the most pervasive and frequent challenges users encounter daily, irrespective of the specific application being used. By prioritizing these categories, we ensure that agents are rigorously tested against the most common sources of disruption. The remaining categories represent critical, albeit less frequent, events. App Malfunction (9.2\%) and Permission Control (9.2\%) are given equal, moderate weight. This reflects that while app crashes and permission dialogs are crucial test cases for the adaptability and recovery capabilities of an agent, they do not occur as constantly as system-level alerts. Finally, UX Disruptions (10.5\%), such as update prompts and rating requests, are included to test an agent's ability to navigate non-critical but workflow-breaking pop-ups, completing the spectrum of realistic interruptions.

Overall, the design of the D-GARA-152 benchmark is two-fold. By concentrating tasks in functionally complex, high-usage applications, it ensures agents are tested in challenging environments. Simultaneously, by distributing interruption types in proportions that realistically model their real-world frequency, it creates a robust and authentic testbed for comprehensively evaluating the adaptability, resilience, and decision-making capabilities of a GUI agent.
\begin{table*}[t] 
    \centering
    \caption{Comparison of different agents benchmarks. D-GARA-152 stands out by supporting both Anomaly detection and Dynamic environments.}
    \label{tab:benchmark_comparison}
    \setlength{\tabcolsep}{4pt} 
    \begin{tabular}{lcccccc}
        \toprule
        \textbf{Benchmark} & \textbf{\#Apps/Web} & \textbf{\#Tasks} & \textbf{Platform} & \textbf{Anomaly?} & \textbf{Dynamic?} & \textbf{Lang.} \\
        \midrule
        Mind2Web & 137 & 2,350 & Desktop Web & \textcolor{red}{$\times$} & \textcolor{red}{$\times$} & EN \\
        GUI Odyssey & 201 & 7,735 & Android Apps & \textcolor{red}{$\times$} & \textcolor{red}{$\times$} & EN \\
        OSWorld & 9 & 369 & Desktop (Apps + Web) & \textcolor{red}{$\times$} & \textcolor{blue}{$\checkmark$} & EN \\
        Windows AgentArena & 11 & 154 & Desktop (Apps + Web) & \textcolor{red}{$\times$} & \textcolor{blue}{$\checkmark$} & EN \\
        WorldGUI & 10 & 315 & Desktop Apps & \textcolor{red}{$\times$} & \textcolor{blue}{$\checkmark$} & EN \\
        Android World & 20 & 116 & Android Apps & \textcolor{red}{$\times$} & \textcolor{blue}{$\checkmark$} & EN \\
        SPA-Bench & 58 & 340 & Android Apps & \textcolor{red}{$\times$} & \textcolor{blue}{$\checkmark$} & EN \& CN \\
        GUI-Robust & 392 & 5,318 & Desktop (Apps + Web) & \textcolor{blue}{$\checkmark$} & \textcolor{red}{$\times$} & EN \& CN \\
        \midrule
        \textbf{D-GARA-152 (ours)} & \textbf{7} & \textbf{152} & \textbf{Android Apps} & \textcolor{blue}{$\checkmark$} & \textcolor{blue}{$\checkmark$} & \textbf{EN \& CN} \\
        \bottomrule
    \end{tabular}
\end{table*}

\section*{Appendix B: Details of Manual Benchmark Collection Pipeline}
The D-GARA framework empowers researchers to construct customized benchmarks for evaluating GUI agent robustness against specific, user-defined anomalies. The workflow is as follows:

Step 1: Anomaly Design and Implementation.
The researcher first identifies and defines the types of real-world interruptions or anomalies to be studied (e.g., incoming call notifications, low-battery warnings, pop-up advertisements). These conceptual anomalies are first implemented as predefined, triggerable events within ADB Smart Test, a specialized Android application (.apk). This step, which leverages native Android development capabilities, prepares each anomaly to be programmatically injected by the main framework during the evaluation pipeline.

Step 2: Task Definition.
Next, the researcher defines a set of meaningful, high-level tasks pertinent to the target application. These tasks are designed to reflect realistic user goals and interaction patterns, such as "book the earliest flight to a destination" or "find and play a specific video playlist." Each task provides a clear objective for the agent to pursue.

Step 3: Manual Trajectory Collection.
Using the DataCollector tool, a human operator manually performs each defined task on the application. This process captures a "golden" trajectory—a correct and complete sequence of actions and corresponding UI states required to accomplish the task goal under normal conditions. This manually collected trajectory serves as the foundational reference for the subsequent automation steps.

Step 4: Injection and Validation Rule Creation.
With the reference trajectory established, the researcher configures the core logic for the automated evaluation. This involves two key activities:
\begin{itemize}
    \item \textbf{Anomaly Injection Rules:} Researchers define when to inject an anomaly by creating a rule in a configuration file. This rule specifies trigger conditions—such as the appearance of certain keywords or UI elements on the screen—that determine the precise moment for injection (e.g., "inject a pop-up dialog when the text 'login' appears in the XML file")

    \item \textbf{Success Validation Rules:} Researchers delineate task accomplishment by formulating a validation criterion that articulates the desired outcome.  This rule delineates the necessary text or UI components that must be on the end screen, establishing a clear and objective standard for successful completion.
\end{itemize}
Once these steps are completed, the D-GARA pipeline is fully configured to run automated experiments.

\section*{Appendix C: Comparison with Existing Benchmarks}
\label{app:comparison}

Table~\ref{tab:benchmark_comparison} presents a detailed comparison between our proposed D-GARA-152 and other mainstream GUI agent benchmarks. 
Existing benchmarks generally fall into two categories: static datasets (e.g., Mind2Web, GUI Odyssey) that focus on offline traces, and dynamic environments (e.g., OSWorld, Android World) that support interactive execution. 

However, a significant gap remains in the current landscape. Most dynamic benchmarks primarily evaluate agents under idealized conditions, neglecting the critical capability of handling unexpected disruptions. While GUI-Robust explicitly targets robustness, it relies heavily on static contexts. We find that it remains largely static in practice, with fewer than 4\% of cases involving actual anomalies. 

D-GARA-152 bridges this gap by adopting an Android World–inspired dynamic assessment framework while simultaneously integrating systematic anomaly injection. Unlike previous works that treat dynamism and robustness in isolation, D-GARA provides a dynamic, anomaly-rich evaluation adaptable to any app or task. Furthermore, D-GARA-152 supports bilingual evaluation (English and Chinese), offering a more comprehensive and reliable measure of agent robustness across different linguistic contexts compared to predominantly English-only benchmarks.

\end{document}